\pdfoutput=1

\documentclass[11pt]{article}

\usepackage[final]{acl}

\usepackage{times}
\usepackage{latexsym}

\usepackage{booktabs,multirow,tabularx,multirow,float}
\usepackage{amsmath,amssymb}

\usepackage{xcolor,caption,subcaption,graphicx}

\newenvironment{myquote}{\list{}{\leftmargin=0.15in\rightmargin=0.15in}\item[]}%
  {\endlist}

\newcolumntype{q}[1]{>{\setlength{\parindent}{-1em}}p{#1}}

\usepackage[T1]{fontenc}

\usepackage[utf8]{inputenc}

\usepackage{microtype}

\usepackage{inconsolata}

\title{Modelling Commonsense Commonalities with Multi-Facet\\ Concept Embeddings}

\author{Hanane Kteich$^{1}$, Na Li$^{3}$, Usashi Chatterjee$^{2}$, Zied Bouraoui$^{1}$, Steven Schockaert$^{2}$\\
  $^{1}$ CRIL CNRS \& University of Artois, France \quad
  $^{2}$ CardiffNLP, Cardiff University, UK \\
  $^{3}$ University of Shanghai for Science and Technology, China\\
  \texttt{\{kteich,bouraoui\}@cril.fr}   \quad \texttt{\{chatterjee,schockaerts1\}@cardiff.ac.uk} \\
   \texttt{li\_na@usst.edu.cn}\\}

\begin{document}
\maketitle

\begin{abstract}
Concept embeddings offer a practical and efficient mechanism for injecting commonsense knowledge into downstream tasks. Their core purpose is often not to predict the commonsense properties of concepts themselves, but rather to identify commonalities, i.e.\ sets of concepts which share some property of interest. Such commonalities are the basis for inductive generalisation, hence high-quality concept embeddings can make learning easier and more robust. Unfortunately, standard embeddings primarily reflect basic taxonomic categories, making them unsuitable for finding commonalities that refer to more specific aspects (e.g.\ the colour of objects or the materials they are made of). In this paper, we address this limitation by explicitly modelling the different facets of interest when learning concept embeddings. We show that this leads to embeddings which capture a more diverse range of commonsense properties, and consistently improves results in downstream tasks such as ultra-fine entity typing and ontology completion.
\end{abstract}

\section{Introduction}
Many knowledge engineering tasks require knowledge about the meaning of concepts. As a motivating example, let us consider the problem of ontology expansion, which consists in uncovering properties of, and relationships between concepts, given the names of these concepts and an initial knowledge base. Despite the popularity of Large Language Models (LLMs), the use of concept embeddings remains attractive in such settings \cite{DBLP:conf/sigir/VedulaNADS018,DBLP:conf/semweb/LiBS19,DBLP:conf/pkdd/MalandriMMN21,DBLP:conf/esws/ShiCDKLZWH23}. Indeed, using LLMs directly is often impractical and highly inefficient, as ontologies can involve tens of thousands of concepts. Concept embeddings can also be integrated with structural features more easily \cite{DBLP:conf/semweb/LiBS19}, for instance by using them to initialise the node embeddings of a Graph Neural Network (GNN). Concept embeddings similarly play an important role in many multi-label classification tasks, especially in the zero-shot or few-shot regime \cite{DBLP:conf/nips/XingROP19,DBLP:conf/mir/YanBWJS21,luo-etal-2021-dont,huang-etal-2022-unified,ma-etal-2022-label}. As a representative example of such a task, we will consider the problem of ultra-fine entity typing \cite{choi-etal-2018-ultra}, which consists in assigning semantic types to mentions of entities, where a total number of around 10K candidate types are considered. In such tasks, the role of pre-trained concept embeddings is to inject prior knowledge about the meaning of the type labels \cite{xiong-etal-2019-imposing,li-etal-2023-ultra}. Note that we cannot straightforwardly accomplish this with LLMs, as they have been found to struggle with information extraction tasks \cite{DBLP:journals/corr/abs-2305-14450}. Moreover, scalability is often an important concern for information extraction systems, which further complicates the use of LLMs. 

We take the view that concept embeddings, in the aforementioned applications, are primarily needed to capture commonalities among the concepts involved. For ontology expansion, this is true by definition, since the task explicitly involves identifying sets of concepts that have some property in common. For ultra-fine entity typing, \citet{li-etal-2023-ultra} reported that directly using pre-trained label embeddings was challenging. Instead, they proposed to cluster the set of labels based on pre-trained concept embeddings, and to use the resulting clusters to structure the label space. The idea of using embeddings to structure the label space also lies at the heart of many traditional approaches for zero-shot and few-shot classification. 

A key limitation of traditional concept embeddings comes from the fact that they primarily reflect basic taxonomic categories. For instance, the embedding of \emph{banana} is typically similar to that of other fruits, but dissimilar from the embeddings of other yellow things.  Some authors have proposed to learn multi-facet embeddings as a way of alleviating these concerns \cite{rothe-schutze-2016-word,jain-etal-2018-learning,alshaikh-etal-2019-learning,alshaikh-etal-2020-mixture}. Essentially, rather than learning a single vector representation of each concept (or entity), they learn a fixed number of different vectors, each focusing on a different facet. However, learning such representations is challenging for two main reasons. First, learning multi-facet representations requires some kind of supervision signal about the facets of interest \cite{DBLP:conf/icml/LocatelloBLRGSB19}, which is not readily available for many domains. Second, existing approaches consider a fixed set of facets, which makes them unsuitable for open-domain settings. Indeed, the facets of interest strongly depend on the nature of the concepts involved. When modelling food, we may be interested in embeddings that capture their nutritional content. When modelling household appliances, we may want a representation that captures where in the house they are typically found. Rather than using the same set of facets for all concepts, we thus need a more dynamic representation framework.

In this paper, we propose a novel method for learning multi-facet concept embeddings based on two key ideas. First, we rely on ChatGPT\footnote{\url{https://openai.com}} to collect a diverse set of (property, facet) pairs, such as (\emph{yellow}, \emph{colour}), (\emph{found in the kitchen}, \emph{location}) or (\emph{sweet}, \emph{taste}), allowing us to treat the problem of learning multi-facet embeddings as a supervised learning problem.  Second, rather than learning several independent vector representations, we only learn a single embedding for each concept, treating facets instead as masks on the set of coordinates. This approach offers several modelling advantages, including the fact that facets can have a hierarchical structure (e.g.\ \emph{colour} is a sub-facet of \emph{appearance}) and the fact that we do not have to tune the number and dimension of the facets a priori. Specifically, we train three BERT \cite{devlin-etal-2019-bert} encoders: one encoder to map concepts onto their embedding, one encoder to map properties onto their embedding, and one encoder to map properties onto the embedding of the corresponding facet. We show that these encoders can be effectively trained using only training data obtained from ChatGPT, although the best results are obtained by augmenting this training data with examples from ConceptNet\footnote{\url{https://conceptnet.io}}. 

\section{Related Work}
\paragraph{Concept Embeddings} The idea that language models of the BERT family can be used for learning concept embeddings has been studied extensively. Some approaches simply use the name of the concept as input to the BERT encoded, possibly together with a short prompt \cite{bommasani-etal-2020-interpreting,vulic-etal-2021-lexfit,liu-etal-2021-fast,gajbhiye-etal-2022-modelling}. Other approaches instead use contextualised representations from sentences mentioning the concept, selected from some corpus \cite{ethayarajh-2019-contextual,bommasani-etal-2020-interpreting,vulic-etal-2020-probing,liu-etal-2021-mirrorwic,DBLP:conf/sigir/LiKBS23}. These approaches have been developed with different motivations in mind. One common motivation is to learn something about the language model itself by inspecting the resulting concept embeddings, such as biases \cite{bommasani-etal-2020-interpreting} or the model's grasp of lexical semantics \cite{ethayarajh-2019-contextual,vulic-etal-2020-probing}. Other authors have rather focused on the use of embeddings for predicting semantic properties of concepts \cite{gajbhiye-etal-2022-modelling,DBLP:conf/sigir/LiKBS23,DBLP:journals/nle/RosenfeldE23}. Our paper can be seen as a continuation of this latter research line, where we aim to improve the range of properties that can be captured by concept embeddings through the use of facet embeddings.

\paragraph{Commonalities} \citet{gajbhiye-etal-2023-deck} recently argued that the main purpose of concept embeddings, when it comes to downstream applications, is usually to identify what different concepts have in common. Specifically, given a set of concepts, they first used the corresponding concept embeddings to predict a set of properties for each concept. The resulting predictions were then filtered using a Natural Language Inference (NLI) model. Finally, properties that were found for at least two concepts where identified as shared properties. They showed, on the task of ultra-fine entity typing, that by augmenting the training data with these shared properties, models were able to generalise better. This idea also relates to the notion of conceptualisation \cite{DBLP:journals/corr/abs-2206-01532,wang-etal-2023-cat}. Essentially, the latter works have suggested to augment commonsense knowledge graphs by generalising the concepts involved. This often involves replacing a specific concept (e.g.\ a football game) by a description referring to some salient property (e.g.\ a relaxing event). \citet{wang-etal-2023-car} showed that the resulting generalisations of commonsense knowledge graphs were useful for zero-shot commonsense question answering. The aforementioned methods all rely on the availability of a set of properties (or hypernyms) that can be used to generalise a given set of concepts. In practice, however, it is hard to obtain comprehensive property sets, which means that many commonalities may not be discovered. Moreover, certain commonalities are hard to describe, even though they intuitively make sense.\footnote{As a simple toy example, among the set $\{\textit{cat},\allowbreak \textit{dog},\allowbreak \textit{goldfish},\allowbreak \textit{rabbit}\}$ the concepts \textit{cat} and \textit{dog} stand out as similar, even though they are not the only pets nor the only mammals.} To avoid such limitations, we identify commonalities by clustering concept embeddings.

\paragraph{Multi-Facet Embeddings}
The idea of capturing different facets of meaning has been studied in the context of disentangled representation learning, especially in computer vision \cite{DBLP:conf/nips/ChenCDHSSA16,DBLP:conf/iclr/HigginsMPBGBML17,DBLP:conf/icml/KimM18,DBLP:conf/nips/ChenLGD18}. When it comes to learning disentangled representation of text, \citet{he-etal-2017-unsupervised} proposed a method for learning aspect embeddings in the context of sentiment analysis, whereas several authors have proposed multi-facet document embeddings \cite{jain-etal-2018-learning,risch-etal-2021-multifaceted,DBLP:conf/webi/KohlmeyerRK21}. \citet{rothe-schutze-2016-word} suggested that word embeddings could be decomposed into meaningful subspaces, which essentially correspond to facets. Most similar to our work, \citet{alshaikh-etal-2019-learning} proposed a method for decomposing a domain-specific concept embedding space into subspaces capturing different facets. To find this decomposition, they relied on the assumption that properties belonging to the same facet tend to have similar word embeddings. Finally, \citet{alshaikh-etal-2020-mixture} proposed a mixture-of-experts model to learn multi-facet concept embeddings directly, using a variant of GloVe \cite{pennington-etal-2014-glove}.

\section{Proposed Approach}
We propose a bi-encoder based concept embedding model which is capable of representing concepts w.r.t.\ different facets. The key stumbling block in previous work on learning multi-facet embeddings has been the difficulty in acquiring a meaningful supervision signal about which properties belong to the same facet (e.g.\  that \emph{large} and \emph{small} both refer to \emph{size}). As explained in Section \ref{secObtainingTrainingData}, we can now overcome this difficulty by collecting property-facet pairs from LLMs. Our proposed model itself is described in Section \ref{secModelFormulation}. Finally, we explain how facet-specific embeddings can be extracted once the model has been trained (Section \ref{secExtractingFacets}).

\subsection{Obtaining Training Data}\label{secObtainingTrainingData}
We need two types of training examples for our model: concept-property judgements (e.g.\ \emph{banana} has the property \emph{rich in potassium}) and property-facet judgments (e.g.\ \emph{rich in potassium} refers to \emph{nutritional content}). We use two sources for obtaining these examples: ChatGPT and ConceptNet.

\paragraph{ChatGPT}
We use the dataset of 109K concept-property judgments that were obtained from ChatGPT by \citet{chatterjee-etal-2023-cabbage}.\footnote{Dataset available from \url{https://github.com/ExperimentsLLM/EMNLP2023_PotentialOfLLM_LearningConceptualSpace}.} To obtain property-facet pairs, we proceed in a similar way, although obtaining suitable information about facets turned out to be more challenging. We obtained the best results with the following prompt, which does not ask about facets explicitly. Instead, we ask about concept-property pairs, but use a format which requires the model to specify the facet of each property that is generated:
\begin{myquote}
\textit{I am interested in knowing common properties that are satisfied by different concepts.
\textbf{1. Sound: loud - thunder, jet engine, siren
2. Temperature: cold - ice, refrigerator, Antarctica 3. Colour: orange - mandarin, basketball, clownfish
4. Shape: round - sun, orange, ball
5. Purpose: used for cleaning - broom, lemon, soap
6. Location: located in the ocean - sand, whale, corals. Please provide me with a list of 30 such examples.}}
\end{myquote}
We used this request with the same prompt 10 times. After this, we change the examples that are given (shown in bold above) and repeat. We manually processed the responses to standardise facet spellings and removed duplicates. For instance, facets were sometimes generated in plural (e.g.\ \emph{colors} rather than \emph{color}), or the same facet was generated with different spellings (e.g.\ \emph{color} and \emph{colour}). Even when changing the examples in the prompt after every 10 requests, the number of unique facet-property pairs that were generated saturated relatively quickly. In total, we obtained 828 unique facet-property pairs, covering 127 unique facets. 

\paragraph{ConceptNet}
Starting from a ConceptNet 5 dump\footnote{\url{https://github.com/commonsense/conceptnet5/wiki/Downloads}}, we first selected the English language triples. Given a triple such as (\textit{boat}, \textit{at location}, \textit{sea}) we create a corresponding concept-property pair (\textit{boat}, \textit{at location sea}) and a property-facet pair (\textit{at location sea}, \textit{at location}). In other words, the ConceptNet relations are treated as facets, and properties are obtained by combining a relation with a tail concept. Not all ConceptNet relations are suitable for this purpose. We specifically used: RelatedTo, FormOf, IsA, UsedFor, AtLocation, CapableOf, HasProperty, HasA, InstanceOf and MadeOf. Furthermore, when creating properties, we only consider tail concepts that appear at least 10 times. We thus end up with 884 distinct properties, 10 facets, 18505 concepts, 884 property-facet pairs and 36955 concept-property pairs.

%
\subsection{Model Formulation}\label{secModelFormulation}
Let us write $\mathcal{D}_{\mathsf{cp}}$ for the set of (concept, property) pairs that are available for training. Similarly, we write $\mathcal{D}_{\mathsf{pf}}$ for the set of available (property, facet) pairs. We build on the following bi-encoder loss from  \cite{gajbhiye-etal-2022-modelling}:
\begin{align*}
\mathcal{L} \, {=} &  -\sum_{(c,p)\in \mathcal{D}_{\mathsf{cp}}} \log \sigma\big(\mathsf{Con}(c)\cdot \mathsf{Prop}(p)\big)\\
&- \sum_{(c,p)\in \mathcal{N}_{\mathsf{cp}}} \log\big(1- \sigma\big(\mathsf{Con}(c)\cdot \mathsf{Prop}(p)\big)\big)
\end{align*}
where $\mathcal{N}_{\mathsf{cp}}$ is a set of negative examples. Specifically, for each positive example $(c,p)$, five negative examples $(c,p')$ are obtained by replacing $p$ by another property $p'$. The concept embedding $\mathsf{Con}(c)$ and property embedding $\mathsf{Prop}(p)$ are obtained by two separate BERT encoders. The concept encoder $\mathsf{Con}$ uses a prompt of the form \texttt{<Concept> means [MASK]}. The property encoder $\mathsf{Prop}$ uses the same prompt. In both cases, the embeddings are obtained from the final-layer embedding of the [MASK] token. However, the concept embedding $\mathsf{Con}(c)$ is normalised (w.r.t.\ the Euclidean norm) whereas the property embedding $\mathsf{Prop}(p)$ is not.

Previous work on multi-facet embeddings has relied on learning multiple concept embeddings, where each concept embedding only captures a subset of all properties \cite{alshaikh-etal-2020-mixture}. This approach has a number of drawbacks, however. For instance, it relies on the idea that each facet is represented using the same number of dimensions, implicitly assumes that the overall number of facets is relatively small, and that the facets are independent from each other. This is particularly problematic in open-domain settings, where a wide range of facets may need to be considered, certain facets only make sense for some concepts (e.g.\ nutritional value only makes sense for food) and facets often have a hierarchical structure (e.g.\ colour is a sub-facet of appearance). Therefore, instead of learning multiple concept embeddings, we instead interpret facets as masks on concept embeddings. 

Specifically, we train a third BERT encoder, $\mathsf{Facet}$, which also takes the property $p$ as input and again uses the same prompt. The idea is that $\mathsf{Facet}(p)$ indicates which coordinates of the concept embeddings are most relevant when modelling the property $p$. 
We define the masked embedding of concept $c$ w.r.t.\ some property $p$ as follows:
\begin{align*}
\mathsf{MC}(c,p) = \frac{\mathsf{Con}(c)\odot \mathsf{Facet}(p)}{\|\mathsf{Con}(c)\odot \mathsf{Facet}(p)\|}
\end{align*}
where we write $\odot$ for the component-wise product. We essentially keep the same bi-encoder model but instead rely on these masked concept embeddings:
\begin{align*}
\mathcal{L}_1\, {=} &  -\sum_{(c,p)\in \mathcal{D}_{\mathsf{cp}}} \log \sigma\big(\mathsf{MC}(c,p)\cdot \mathsf{Prop}(p)\big)\\
& - \sum_{(c,p)\in \mathcal{N}_{\mathsf{cp}}} \log\big(1- \sigma\big(\mathsf{MC}(c,p)\cdot \mathsf{Prop}(p)\big)\big)
\end{align*}
Without further supervision, the facet encoder does not learn meaningful facets. Therefore, we use the (property, facet) examples from $\mathcal{D}_{\mathsf{pf}}$ to ensure that properties which belong to the same facet have a similar facet embedding. For a given facet $f$, we write $\mathcal{P}_f$ for the set of properties that we know to belong to this facet, i.e.\ $\mathcal{P}_f = \{ p\,|\, (p,f)\in \mathcal{D}_{\mathsf{pf}}\}$. We use the  InfoNCE loss: 
\begin{align*}
\mathcal{L}_2\, {=} &
-\sum_f \sum_{p,q\in \mathcal{P}_f} \log \frac{\exp\left(\frac{\cos(\mathsf{F}(p),\mathsf{F}(q))}{\tau}\right)}{\sum\exp\left(\frac{\cos(\mathsf{F}(p),\mathsf{F}(r))}{\tau}\right)}
\end{align*}
where we abbreviated $\mathsf{Facet}(p)$ as $\mathsf{F}(p)$, the sum in the denominator ranges over $r\in \{q\}\cup \{p\,|\, (p,f)\notin \mathcal{D}_{\mathsf{pf}}\}$, and the temperature $\tau>0$ is a hyperparameter. The InfoNCE loss encourages properties which belong to the same facet to have facet embeddings that are more similar to each other than to the facet embeddings of properties which do not. The overall model is trained by optimising the loss $\mathcal{L}_1 + \mathcal{L}_2$.

\subsection{Extracting Facet-Specific Representations}\label{secExtractingFacets}
The model from Section \ref{secModelFormulation} can be used in several ways. First, we can simply use the concept embeddings $\mathsf{Con}(c)$ to represent each concept $c$. In this case, the purpose of having facets is to ensure that the concept embeddings capture a broader range of properties, but we only consider these facets during training. We will refer to this approach as \textbf{ConEmb-F}. The concept embeddings from the standard bi-encoder, without facet embeddings, will be referred to as \textbf{ConEmb}. 

In some applications, concept embeddings are used for clustering concepts. The purpose of multi-facet embeddings is to ensure that different kinds of clusters can be found. In such settings, we extract different facet-specific concept embeddings from the model. Specifically, let $\mathcal{P}=\{p_1,...,p_m\}$ be the set of properties of interest. For each property $p_i$ we have a corresponding facet vector $\mathbf{f}_i = \mathsf{Facet}(p_i)$. We use the K-means algorithm to cluster these facet vectors into clusters $\mathcal{X}_1,...\mathcal{X}_k$ and treat each of these clusters as facet. 
We associate each concept $c$ with $k$  facet-specific representations $\mathbf{c}_1,...,\mathbf{c}_k$, defined as:
\begin{align}\label{eqFacetSpecificEmbeddingComputation}
\mathbf{c}_j = \frac{\mathsf{Con}(c) \odot \big(\sum_{p_i\in \mathcal{X}_j}\mathbf{f}_i\big)}{\big\|\mathsf{Con}(c) \odot \big(\sum_{p_i\in \mathcal{X}_j}\mathbf{f}_i\big)\big\|}
\end{align}
The representations obtained by this approach depend on how the set of properties $\mathcal{P}$ is chosen. For our experiments, we simply set $\mathcal{P}$ to be the set of all properties that appear in our training set. 


\section{Experiments}
We analyse the effectiveness of the proposed multi-facet concept embedding model. We intrinsically evaluate the embeddings on predicting commonsense properties (Section \ref{secPredictingProperties}) and outlier detection (Section \ref{secOutlierDetection}). We furthermore consider two downstream applications: ontology completion (Section \ref{secOntologyCompletion}) and ultra-fine entity typing (Section \ref{secUFET}).\footnote{Our datasets, pre-trained models and implementation are available at \url{https://github.com/hananekth/facets_concepts_embeddings}}


\begin{table*}[t]
\centering
\footnotesize
\begin{tabular}{lcll ccc ccc}
\toprule
&\textbf{LM} & \textbf{Train properties ($\mathcal{D}_{\mathsf{cp}}$)} & \textbf{Train facets ($\mathcal{D}_{\mathsf{pf}}$)} & \multicolumn{3}{c}{\textbf{McRae}} &\multicolumn{3}{c}{\textbf{CSLB}} \\
 \cmidrule(lr){5-7}\cmidrule(lr){8-10}
&&& & \textbf{Con} & \textbf{Prop} & \textbf{C+P} & \textbf{Con} & \textbf{Prop} & \textbf{C+P}  \\
 \midrule
BiEnc$^*$ & BB & MSCG & \multicolumn{1}{c}{-} & 79.8 & 49.6 & 44.5 & 54.5 & 39.1 & 32.6\\
BiEnc$^*$ & BL & MSCG & \multicolumn{1}{c}{-} &80.5 & 49.3 & 45.5 &57.7 & 41.8 & 36.4\\
BiEnc$^*$ & RB & MSCG & \multicolumn{1}{c}{-} &75.6 & 42.4 & 38.1 & 49.9 & 36.4 & 24.3\\
BiEnc$^*$ & RL & MSCG & \multicolumn{1}{c}{-} &80.1 & 46.5 & 42.5 & 59.0 & 42.5 & 36.0\\
\midrule
BiEnc & BL & CN & \multicolumn{1}{c}{-} & 78.0 & 56.7 &	51.8	& 61.4	& 49.6 &	50.0\\
BiEnc & BL & ChatGPT & \multicolumn{1}{c}{-} & 80.5&	57.3&	56.6	&65.1&	56.5&	52.7\\
BiEnc & BL & ChatGPT+CN & \multicolumn{1}{c}{-} & 81.7	& 62.1	& 59.5	& 67.8 &	59.6&	53.1\\
\midrule
BiEnc & BB & ChatGPT+CN & \multicolumn{1}{c}{-} & 76.2	& 60.6	& 58.4	& 66.9&	56.6&	51.8\\
BiEnc & RB & ChatGPT+CN & \multicolumn{1}{c}{-} & 75.8 & 60.1 & 58.2 & 66.1 & 56.3 & 51.7\\
BiEnc & RL & ChatGPT+CN & \multicolumn{1}{c}{-} &  80.8 & 61.7 & 59.3 & 67.2 & 58.8& 52.7\\
\midrule
BiEnc-F & BL & ChatGPT+CN & CN & 84.3 &63.5&	57.7 & 69.4 & 61.0 & 59.9\\
BiEnc-F & BL & ChatGPT+CN & ChatGPT & 84.3 & 64.9	& 65.5 & 69.5 &61.6& 61.9\\
BiEnc-F & BL & ChatGPT+CN & ChatGPT+CN & \textbf{86.2} & \textbf{65.9 }& \textbf{67.0} & \textbf{70.3}& \textbf{63.6}& \textbf{63.0}\\
\midrule
BiEnc-F & BB & ChatGPT+CN & ChatGPT+CN & 82.1 & 63.0 &	61.2 & 65.3 & 60.2 & 59.9\\
BiEnc-F & RB & ChatGPT+CN & ChatGPT+CN & 81.5 & 62.3 & 60.8 & 65.0& 59.6 & 61.3\\
BiEnc-F & RL & ChatGPT+CN & ChatGPT+CN & 85.6 & 65.1 & 65.9 & 69.2 & 63.1 & 62.8\\
\bottomrule
\end{tabular}
\caption{Results for commonsense property prediction in terms of F1 ($\%$). Results marked with $^*$ were taken from \citet{gajbhiye-etal-2022-modelling}. MSCG corresponds to the training set from \citet{gajbhiye-etal-2022-modelling}; ChatGPT and CN (ConceptNet) refer to the training sets that were described in Section \ref{secObtainingTrainingData}. We evaluate using BERT-base-uncased (BB), BERT-large-uncased (BL), RoBERTa-base (RB) and RoBERTa-large (RL). \label{tabPropertyPrediction}}
\end{table*}

\subsection{Predicting Commonsense Properties}\label{secPredictingProperties}
The use of facets should lead to concept embeddings that capture a wider range of properties. To test this hypothesis, we consider the task of commonsense property prediction, which we treat as a binary classification problem: given a concept and a commonsense property, decide whether the property is satisfied by the concept or not. The difficulty of this task depends on how the training-test split is constructed. One strategy, called \textbf{concept split}, ensures that the concepts appearing in the training and test data are different, but the properties are not. \citet{gajbhiye-etal-2022-modelling} found that simple nearest neighbour strategies can do well on this task, meaning that this variant does not adequately assess whether the concept embeddings capture commonsense knowledge. For this reason, they proposed a \textbf{property split}, where the properties appearing in training and test are different, but the concepts are the same. Finally, they also considered a  \textbf{C+P} (concept+property) split, where both the concepts and properties are different in training and test. In all cases, we first pre-train the encoders $\mathsf{Con}$, $\mathsf{Prop}$ and $\mathsf{Facet}$ on the ChatGPT and/or ConceptNet training data (see Section \ref{secObtainingTrainingData}), before fine-tuning on the training splits of the property prediction benchmarks.\footnote{Detailed training details can be found in the appendix.} 

Table \ref{tabPropertyPrediction} summarises the results for all three settings, using two standard benchmarks for commonsense property prediction: the extension of the McRae property norms dataset \cite{mcrae2005semantic} that was introduced by \citet{DBLP:conf/cogsci/ForbesHC19} and the augmented version of CSLB\footnote{\url{https://cslb.psychol.cam.ac.uk/propnorms}} introduced by \citet{DBLP:journals/corr/abs-2205-06910}. Our main baseline is the bi-encoder model from \citet{gajbhiye-etal-2022-modelling}, shown as \textbf{BiEnc}, which also forms the basis for our facet-based model. We report the results from \citet{gajbhiye-etal-2022-modelling}, which are for a model that was pre-trained on Microsoft Concept Graph \cite{DBLP:journals/dint/JiWSZWY19} and GenericsKB \cite{DBLP:journals/corr/abs-2005-00660}, as well as results for models that we trained on the ChatGPT and ConceptNet training sets (see Section \ref{secObtainingTrainingData}). Finally, we show the result of our full model (i.e.\ with the facet encoder), shown as \textbf{BiEnc-F}. We compare four different encoders: BERT-base, BERT-large, RoBERTa-base and RoBERTa-large. 

The results clearly show the benefit of using facets, as the BiEnc-F models consistently and substantially outperform the BiEnc baselines. Comparing the different training sets, the ChatGPT training examples are more effective than the ConceptNet examples, but the best results are obtained when both sources are combined. Among the different LMs, BERT-large achieves the best results.

Based on the results from this experiment, for the remaining experiments, we will focus on the model based on BERT-large, which is trained on both the ChatGPT and ConceptNet training examples.

\begin{table}[t]
\centering
\footnotesize
\setlength\tabcolsep{2.5pt}
\begin{tabular}{lccc}
\toprule
 & \textbf{ConEmb} & \textbf{ConEmb-F} & \textbf{MultiConEmb}\\
\midrule
Dangerous & 9	& 3	& \textbf{13} \\
Edible & 23	& 26	& \textbf{67} \\
Flies & 23	& 34	& \textbf{77}\\
Hot &7	&11	& \textbf{40}\\
Lives in water & 33	& 48	& \textbf{83}\\
Produces noise & 13	& 13	& \textbf{48}\\
Sharp & 33	& 30	& \textbf{67}\\
Used by children &8	& 9	& \textbf{17} \\
Used for cooking& 59 &	53	& \textbf{93} \\
Worn on feet &18	& 13	& \textbf{54}\\
\bottomrule
\end{tabular}
\caption{Results for outlier detection (exact match $\%$). \label{tabOutlierDetection}}
\end{table}

\subsection{Outlier Detection}\label{secOutlierDetection}
To evaluate whether our facet-based concept embeddings can help us to identify commonalities, we consider the task of outlier detection \cite{camacho-collados-navigli-2016-find,DBLP:conf/iclr/BlairMB17,brink-andersen-etal-2020-one}. In each instance of this task, we are given a set of concepts (or entities). Some of these concepts have a particular property in common, and the task consists in identifying these concepts (without being given any information about the shared property itself). This task has traditionally been used as an intrinsic benchmark for evaluating word embeddings. 

\paragraph{Dataset} Existing benchmarks mostly focus on broad taxonomic categories, whereas we are specifically interested in identifying shared commonsense properties. We therefore constructed a new outlier detection benchmark based on the extended McRae dataset from \citet{DBLP:conf/cogsci/ForbesHC19}. 
To create an outlier detection problem instance, we first select a property from this dataset (e.g.\ \emph{dangerous}) as well as 3 concepts which have this property and 7 outlier concepts which do not. Many of the properties in the McRae dataset are not suitable for our benchmark, either because they correspond to taxonomic categories (e.g.\ \emph{an animal} or \emph{edible}) or because they are too ambiguous or noisy (e.g.\ \emph{accordion}, \emph{car} and \emph{escalator} are described as having the property \emph{fun}, but \emph{airplane} is not). Therefore, we manually selected 10 properties which do not suffer from these limitations. For each property, we manually clustered the concepts with this property into broad taxonomic groups.\footnote{The resulting clusters can be found in the appendix.} When selecting the 3 positive concepts, for a given instance, we ensure that all three examples come from a different  group. When selecting the 7 outliers, we check that they do not share any properties. Specifically, we ensure that any two of the outliers do not have any properties in common in the extended McRae dataset, in ConceptNet or in Ascent++\footnote{\url{https://ascentpp.mpi-inf.mpg.de}}. For each property, we sample 100 problem instances, following this process. We report the results in terms of exact match, i.e.\ the prediction for a given instance is labelled as correct if the three positive examples were correctly identified. We report the percentage of correctly labelled instances, for each property.

\paragraph{Methods}
We compare three strategies for detecting outliers. For the method denoted \textbf{ConEmb}, we use the ConEmb embeddings as follows. For each concept $c$, we find the second and third nearest neighbour. Let us denote these as $c_2$ and $c_3$. If $c$ is a positive example, $\cos(\mathsf{Con}(c),\mathsf{Con}(c_2))$ should be high and $\cos(\mathsf{Con}(c),\mathsf{Con}(c_3)$ should be low. We thus score each concept as
$\textit{score}(c) = \cos(\mathsf{Con}(c),\mathsf{Con}(c_2)) - \cos(\mathsf{Con}(c),\mathsf{Con}(c_3))$.
As positive examples, we then select the concept with the highest score along with its two nearest neighbours. The method denoted \textbf{ConEmb-F} uses the same strategy, but instead uses the ConEmb-F embeddings. Finally, when using the method denoted \textbf{MultiConEmb}, we first obtain 10 facet-specific embeddings of each concept, using \eqref{eqFacetSpecificEmbeddingComputation}. We then first apply the same method as before to each of the 10 facet-specific embedding spaces. Finally, we select the prediction for the facet where the score of the highest-scoring concept was maximal.

\paragraph{Results}
The results are summarised in Table \ref{tabOutlierDetection}. MultiConEMb, which exploits facet-specific representations, substantially outperforms the baselines. The performance of ConEmb and ConEmb-F is comparable, which is as expected: even though ComEmb-F was trained using facets, this method represents concepts as single vectors, and the similarities between these concept vectors still mostly reflect taxonomic relatedness.

\begin{table}
\centering
\footnotesize
 \setlength\tabcolsep{3pt}
\begin{tabular}{l ccccc}
\toprule
& \textbf{Wine} & \textbf{Econ} & \textbf{Olym} & \textbf{Tran} & \textbf{SUMO}\\
\midrule
GloVe$^*$ & 14.2 & 14.1 & 9.9 & 8.3 & 34.9 \\
Skipgram$^*$ & 13.8 &  13.5 & 8.3 & 7.2 & 33.4\\ 
Numberbatch$^*$ & 25.6 &26.2  &26.8& 16.0 &47.3\\
MirrorBERT$^*$ & 22.5 & 23.8 & 20.9& 12.7 & 40.1 \\
MirrorWiC$^*$ & 24.7& 24.9 & 22.1 & 13.9& 46.9 \\
ConCN$^*$ & 31.3 & 32.4 & 29.7 & 20.9 & 52.6\\
\midrule
ConEmb  & 30.8 & 30.5 & 28.6 & 19.8 & 51.3 \\
ConEmb-F & 31.2 &31.8 & 30.4 & 20.9 & 51.7 \\
\midrule
Clu (ConCN) & 35.3 & 33.1 & 32.5 & 21.6 & 52.2 \\
Clu (ConEmb-F) & 36.9 & 34.2 & \textbf{34.6} & 22.1 & 53.3 \\
MClu (ConCN) & 39.8 & 35.9 & 32.6 & 22.7 & 54.2 \\
MClu (ConEmb-F) & \textbf{39.9} & \textbf{36.3} & 32.9 & \textbf{23.1} & \textbf{55.4} \\
\bottomrule
\end{tabular}
\caption{Results for ontology completion in terms of F1 ($\%$). Results marked with $^*$ were taken from \citet{DBLP:conf/sigir/LiKBS23}. ConEmb and ConEmb-F were trained using both the ChatGPT and ConceptNet training sets.\label{tabOntologyCompletionResults}}
\end{table}

\subsection{Ontology Completion}\label{secOntologyCompletion}
Ontologies use rules to encode how the concepts of a given domain are related. They generalise taxonomies by allowing the use of logical connectives to encode these relationships. \citet{DBLP:conf/semweb/LiBS19} introduced a framework for predicting missing rules in ontologies using a Graph Neural Network. The nodes of the considered graph correspond to the concepts from the ontology, and the input representations are pre-trained concept embeddings. Recent work has used this model to evaluate concept embeddings, as its overall performance is sensitive to the quality of the input representations \cite{DBLP:conf/sigir/LiKBS23}. The intuition underpinning the model is closely aligned with the idea of modelling concept commonalities. Essentially, if the ontology contains the rules\footnote{In description logic syntax, $X\sqsubseteq Y$ means that every instance of the concept $X$ is also an instance of the concept $Y$, i.e.\ it represents the rule ``if $X$ then $Y$''.} $X_1 \sqsubseteq Y,...,X_k \sqsubseteq Y$ and we know from the pre-trained embeddings that $X_{k+1}$ is similar to $X_1,...,X_k$ then it is plausible that the rule $X_{k+1}\sqsubseteq Y$ is valid as well.

We test the effectiveness of our model in two ways. First, we use the \textbf{ConEmb-F} concept embeddings as input features, which allows for a direct comparison with the effectiveness of other concept embedding models. In this case, the use of facets only affects how the concept embeddings are learned. As a second strategy, referred to as \textbf{MClu}, we first obtain 10 facet-specific embeddings of all the concepts, using \eqref{eqFacetSpecificEmbeddingComputation}. We then cluster the concepts in each of the facet-specific concept embeddings separately. For this step, we have relied on affinity propagation. This results in 10 different clusterings of the concepts. For each cluster $\mathcal{C}$ in each of these clustering we add a fresh concept $Y_{\mathcal{C}}$ to the ontology, and for every concept $X$ in $\mathcal{C}$, we add the rule $X\sqsubseteq Y_{\mathcal{C}}$. We then apply the standard GNN model from \citet{DBLP:conf/semweb/LiBS19} to the resulting extended ontology. As a baseline, we also apply the clustering strategy to the ConEmb concept embeddings, which we refer to as \textbf{Clu}. Note that in this case there is only one clustering. Note that when \textbf{MClu} or \textbf{Clu} are used, we still need to use concept embeddings to use as input features. We show results with the ConCN embeddings from \citet{DBLP:conf/sigir/LiKBS23} and for our ConEmb-F embeddings.

The results in Table \ref{tabOntologyCompletionResults} show that the ConEmb-F input embeddings consistently outperform the ConEmb vectors. Moreover, they achieve a performance which is similar to that of the ConCN embeddings, which is the current state-of-the-art. Moreover, the proposed clustering strategies, which have not previously been considered for ontology completion, are highly effective. The MClu variant outperforms Clu in all but one case, which shows the benefit of explicitly considering facet-specific embeddings. We can also see that the ConEmb-F embeddings as input features perform better than  the ConCN embeddings, when used in combination with the clustering strategies.

\begin{table}
\footnotesize
\centering
\begin{tabular}{lc}
\toprule
 & \textbf{F1} \\
\midrule
Base model$^{\dagger}$ & 49.2\\
Properties$^*$ & 50.9 \\
Clu (ConCN)$^{\dagger}$  & 50.4 \\
Clu (ConEmb)$^*$ & 50.6 \\
Clu (ConEmb-F) & 50.8 \\
Clu (ConCN) + properties$^*$ & 50.9 \\
Clu (ConEmb) + properties$^*$ & 51.1 \\
MClu & \textbf{51.3}\\
\bottomrule
\end{tabular}
\caption{Results for ultra-fine entity typing, using a BERT-base entity encoder with augmented label sets. Results marked with $^*$ were taken from \citet{gajbhiye-etal-2023-deck}; results marked with $^\dagger$ were taken from \citet{li-etal-2023-ultra}. \label{tabUFETresults}}
\end{table}

\subsection{Ultra-Fine Entity Typing}\label{secUFET}
We consider the task of ultra-fine entity typing \cite{choi-etal-2018-ultra}, which was also used by \citet{gajbhiye-etal-2023-deck} to demonstrate the usefulness of modelling concept commonalities. Given a sentence in which an entity mention is highlighted, the task consists in assigning labels that describe the semantic type of the entity. The task is formulated as a multi-label classification problem with around 10K candidate labels. Many of the candidate labels only have a small number of occurrences in the training data. This makes it paramount to rely on some kind of pre-trained knowledge about the meaning of the labels. \citet{li-etal-2023-ultra} proposed a simple but surprisingly effective strategy: use pre-trained concept embeddings to cluster the labels and augment the training set with labels that refer to these clusters. For instance, if a training example is labelled with label $l$ and this label belongs to cluster $c$ then they add the synthetic label ``cluster c'' to this training example. This intuitively teaches the model which labels are semantically related, as the training objective encourages instances which are labelled with ``cluster $c$'' to be linearly separated from other instances. \citet{gajbhiye-etal-2023-deck} improved on this strategy by instead identifying commonsense properties that were satisfied by the different concepts/labels, and by augmenting the training examples with these properties, instead of synthetic cluster labels. This use of shared properties has the advantage that a broader range of commonalities can be identified, whereas clustering standard concept embeddings leads to clusters that only reflect standard taxonomic categories. Our hypothesis is that we can achieve the same benefits by clustering our multi-facet representations, and that the use of clusters can potentially lead us to capture finer-grained commonalities. 

Table \ref{tabUFETresults} summarises the results. All results were obtained using the DenoiseFET model from \citet{DBLP:conf/ijcai/Pan0022}. The \emph{base model} in Table \ref{tabUFETresults} shows the results if we use this model without augmenting the training labels. \emph{Properties} refers to the strategy from \citet{gajbhiye-etal-2023-deck}, which adds labels corresponding to shared properties, while \textbf{Clu} is the strategy from \citet{li-etal-2023-ultra}, which adds labels corresponding to clusters. We show results for this clustering strategy with three different concept embeddings: the ConCN embeddings from \citet{DBLP:conf/sigir/LiKBS23} as well as ConEmb and ConEmb-F. The approach where we use clusterings from different facet-specific embeddings is shown as \textbf{MClu}. We can see that MClu achieves the best results, which confirms the usefulness of facet-specific representations for this task. When using the Clu strategy, ConEmb-F also slighly outperforms ConEmb.

\section{Conclusions}
Many applications rely on background knowledge about the meaning of concepts. What is needed often boils down to knowledge about the commonalities between different concepts, as this forms the basis for inductive generalisation. Clustering pre-trained concept embeddings has been proposed in previous work as a viable strategy for modelling such commonalities. However, the resulting clusters primarily capture taxonomic categories, while commonalities that depend on various commonsense properties are essentially ignored. In this paper, we proposed a simple strategy for obtaining more diverse representations, by taking into account different facets of meaning when training the concept encoder. We found that the resulting concept representations lead to consistently better results, across all the considered tasks.  

\paragraph{Acknowledgments}
This work was supported by EPSRC grants EP/V025961/1 and EP/W003309/1, ANR-22-CE23-0002 ERIANA and HPC resources from GENCI-IDRIS (Grant 2023-[AD011013338R1]).

\section*{Limitations}
Our approach relies on encoders from the BERT family, which are much smaller than recent language models. We did some initial experiments with Llama 2, but were not successful in obtaining better-performing concept embeddings with this model. While it seems likely that future work will reveal more effective strategies for using larger models, our use of BERT still has the advantage that we can efficiently encode a large number of labels, which remains important for applications such as extreme multi-label text classification. 

We have only looked at modelling commonsense properties of concepts. Modelling facets of meaning is intuitively also important for modelling named entities (e.g.\ for entity linking) and for sentence/document embeddings (e.g.\ for retrieval). An analysis of facet-based models for such applications is left as a topic for future work.

\bibliography{custom,anthology}

\appendix

\section{Training Details}

\paragraph{Predicting Commonsense Properties}
We use the AdamW optimizer with learning rate 2e-5. We use early stopping with a patience of 20.
We use the same settings as \citet{gajbhiye-etal-2022-modelling} for both fine-tuning and pre-training. During pre-training, we randomly select 10000 pairs for tuning.  For the concept split we use the  fixed training-test split from \citet{gajbhiye-etal-2022-modelling}. For property split, we use 5-fold cross-validation, while for C+P, we used the 3 $\times$ 3 fold cross-validation strategy. During fine-tuning, for all splits, we randomly select 20\% from the training set as validation data for model selection. We use a batch size of 32 and set the maximal sequence length for the concept and property prompts to 32.

\paragraph{Outlier Detection}
For these experiments, we use the same bi-encoder models as for predicting commonsense properties. Specifically, we have used the model initialised form BERT-large-uncased, which was trained using ConceptNet and ChatGPT. However, in this case, the model is used without further fine-tuning. 

\paragraph{Ontology Completion}
To get concept clusters from the affinity propagation algorithm, we tune the preference values from $\{0.5,0.6,0.7,0.8,0.9\}$. We set the learning rate to 1e-2, the maximum number of epochs to 200 and the dropout rate to 0.5. We use AdamW as optimizer and tune the number of hidden dimensions from $\{8, 16, 32, 64\}$ and the number of GNN layers from $\{3,4,5\}$. We use a weight decay of 5e-2. 

\paragraph{Ultra-Fine Entity Typing}
To get concept clusters, we again use affinity propagation and select the preference values from $\{0.5,0.6,0.7,0.8,0.9\}$. We use a learning rate of 2e-5 with AdamW as optimizer, a batch size of 16 and a maximum number of 500 epochs.





\begin{table*}[t]
\centering
\footnotesize
\setlength\tabcolsep{15pt}
\begin{tabular}{@{}lq{350pt}@{}}
\toprule
\textbf{Property} & \textbf{Positive Examples} \\
\midrule
\multirow{4}{*}{Dangerous} & 
1: alligator, bear, beehive, bull, cheetah, crocodile, lion, rattlesnake, tiger\\
& 2: bazooka, bomb, bullet, crossbow, dagger, grenade, gun, harpoon, missile, pistol, revolver, rifle, rocket, shotgun, sword, axe, baseball bat, knife, machete\\
& 3: motorcycle\\
\midrule
\multirow{8}{*}{Edible} & 1: apple, banana, cranberry, tomato, tangerine, strawberry, spinach, rhubarb, raspberry, radish, pumpkin, prune, mushroom, parsley, walnut, rice, raisin, potato, plum, pineapple, pepper, peas, pear, peach, orange, onions, olive, nectarine, lime, lettuce, lemon, grapefruit, grape, garlic, cucumber, corn, coconut, cherry, celery, cauliflower, carrot, cantaloupe, cabbage, broccoli, blueberry, beets, beans, avocado, asparagus\\
& 2: bread, cake, cheese, hot dog, pizza, sandwich, pie, donut, biscuit\\
& 3: crab, deer, hare, octopus, salmon, turkey, tuna, trout, squid, shrimp, sardine, pig, octopus, lobster, lamb, goat, cow, clam, chicken\\
\midrule
\multirow{5}{*}{Flies} & 1: bird, crow, dove, duck, eagle, falcon, flamingo, goose, hawk, woodpecker, pigeon, owl, peacock, seagull\\
& 2: butterfly, hornet, housefly, wasp, moth\\
& 3: airplane, helicopter, jet, missile, rocket\\
& 4: balloon, kite, frisbee\\
\midrule
\multirow{3}{*}{Hot} & 1: bathtub\\
& 2: cigar, cigarette, candle\\
& 3: hair drier, kettle, oven, stove, toaster\\
\midrule
\multirow{4}{*}{Lives in water} & 1: alligator, crocodile, otter, turtle, seal, frog, flamingo, salamander, swan, walrus\\
& 2: clam, crab, lobster, octopus, squid, shrimp\\
& 3: dolphin, eel, goldfish, whale, salmon, sardine, trout, tuna\\
& 4: boat, sailboat, ship, submarine, yacht, canoe\\
\midrule
\multirow{4}{*}{Produces noise} & 1: accordion, bagpipe, clarinet, flute, harp, piano, trombone, violin\\
& 2: airplane, ambulance, helicopter\\
& 3: bomb, cannon, grenade\\
& 4: cell phone, hair drier, stereo\\
\midrule
\multirow{5}{*}{Sharp} &1: axe, bayonet, dagger, machete, knife, spear, sword \\
& 2: chisel, corkscrew, screwdriver, scissors\\
& 3: blender, grater\\
& 4: razor\\
& 5: pin\\
\midrule
\multirow{7}{*}{Used by children} & 1: balloon\\
& 2: buggy\\
& 3: crayon, paintbrush, pencil\\
& 4: doll, teddy bear, toy\\
& 5: earmuffs\\
& 6: frisbee, kite\\
& 7: skateboard, sled, tricycle\\
\midrule
\multirow{5}{*}{Used for cooking}& 1: apron\\
& 2: blender, grater, mixer\\
& 3: bowl, colander, pan, pot, skillet, strainer\\
& 4: kettle, microwave, oven, stove, toaster\\
& 5: knife, ladle, spatula, spoon, tongs\\
\midrule
\multirow{3}{*}{Worn on feet} & 1: boots, sandals, shoes, slippers\\
& 2: nylons, socks\\
& 3: skis, snowboard\\
\bottomrule
\end{tabular}
\caption{Results for outlier detection in terms of F1 ($\%$). \label{tabOutlierDetectionDataset}}
\end{table*}

\section{Outlier Detection Dataset}
Table \ref{tabOutlierDetectionDataset} shows the properties that were selected from the McRae dataset for constructing the outlier detection benchmark. For each property, we selected concepts that are asserted to have this property in the dataset, and we organised them into broad taxonomic categories, which are also shown in Table \ref{tabOutlierDetectionDataset}.

\section{Qualitative Analysis}\label{secQualitativeAnalysis}
Table \ref{tabQualitativeMcRae} shows examples of the nearest neighbours of \textit{frisbee} and \textit{bureau} in some of the facet-specific embedding spaces that are used by the MClu strategy. For this analysis, we use the set of concepts from the McRae dataset. These examples illustrate how different facets emphasise different aspects. 

For the case of \textit{frisbee}, the first facet links this concept to other sports related terms. In the second facet, it is instead related to round things. In the third facet, the top neighbours are related to kids. Due to the way in which the facets are learned (e.g.\ by considering a fixed number of facet embedding clusters), there are also some facets that reflect a mixture of different aspects. For instance, the last facet in Table \ref{tabQualitativeMcRae} combines elements of the first three facets, i.e.\ the nearest neighbours cover sports related concepts, kids related concepts, and round things. For \textit{bureau}, in the first facet we find office related terms. In the second facet, the top neighbours are different types of furniture. In facets 3 and 4 we see a mixture of different kinds of terms.

Table \ref{tabQualitativeUFET} similarly shows examples of nearest neighbours in different facets for the label space from the UFET dataset. 

\begin{table*}[t]
\footnotesize
\centering
\begin{tabular}{lp{380pt}}
\toprule
\textbf{Concept} & \textbf{Neighbours}\\
\midrule
\multirow{4}{*}{frisbee} & \textit{Facet 1:} tricycle, surfboard, sports\_ball, tennis\_racket, snowboard, kite, doll, balloon, toy, pie\\
& \textit{Facet 2:} balloon, pie, sports\_ball, cake, donut, teddy\_bear, surfboard, kite, doll, toy\\ 
& \textit{Facet 3:}  kite, doll, balloon, toy, tricycle, moth, pie, sports\_ball, surfboard, football\\ 
& \textit{Facet 4:}  tricycle, surfboard, sports\_ball, tennis\_racket, snowboard, kite, doll, balloon, toy, pie\\
\midrule
\multirow{4}{*}{bureau} & \textit{Facet 1:} envelope, certificate, typewriter, doorknob, fence, cabinet, carpet, shelves, bookcase, gopher\\
& \textit{Facet 2:} desk, dining\_table, table, shelves, bookcase, envelope, typewriter, gopher, escalator, peg\\
& \textit{Facet 3:} shelves, bookcase, envelope, desk, peg, gopher, tack, cabinet, handbag, hook\\ 
& \textit{Facet 4:} bookcase, cabinet, shelves, desk, doorknob, dining\_table, envelope, typewriter, handbag, lamp\\ 
\bottomrule
\end{tabular}
\caption{Nearest neighbours for different facets, using the concepts from the McRae dataset.\label{tabQualitativeMcRae}}
\end{table*}

\begin{table*}[t]
\footnotesize
\centering
\begin{tabular}{lp{370pt}}
\toprule
\textbf{Concept} & \textbf{Neighbours}\\
\midrule
keyboard & \textit{Facet 1:} bass, rhythm guitar, lead guitar, drum major, pianist, bass drum, bass guitar, bassist, bass guitarist, air guitar \\
\cmidrule{2-2}
& \textit{Facet 2:} processor, organ, desktop, mac, storage, file system, thumb drive, file extension, touch screen, desktop environment \\
\midrule
business card & \textit{Facet 1:} flash card, smart card, bank card, graphics card, green card, index card, network card, phone card, press card, punch card, report card, trade card\\
\cmidrule{2-2}
& \textit{Facet 2:} business class, business cycle, business day, business economics, business end, business ethics, business intelligence, business logic\\ 
\bottomrule
\end{tabular}
\caption{Nearest neighbours for different facets, using the labels from the UFET dataset.\label{tabQualitativeUFET}}
\end{table*}

\end{document}